%% file: steering-style-topic.tex
\documentclass[11pt,letterpaper]{article}
\usepackage{emnlp2017}
\usepackage{times}
\usepackage{latexsym}
\usepackage{url}
\usepackage{hyperref}
\usepackage{enumitem}
\usepackage{amsmath}
\usepackage{amsfonts}
\usepackage{svg}
\usepackage{booktabs}
\usepackage{multirow}
\usepackage{xspace}
\usepackage{tabularx}
\usepackage{tabulary}
\usepackage{multirow}

\input{commands.tex}

\emnlpfinalcopy{}

\def\emnlppaperid{1380}

\title{Steering Output Style and Topic in Neural Response Generation}

\author{
	Di Wang$^{1,2}$, Nebojsa Jojic$^1$, Chris Brockett$^1$, and Eric Nyberg$^2$\\
	$^1$Microsoft Research, Redmond, WA, USA \\
    $^2$School of Computer Science, Carnegie Mellon University, Pittsburgh, PA, USA\\
    {\tt diwang@cs.cmu.edu}, {\tt jojic@microsoft.com}
	}
\date{}

\begin{document}
\maketitle

\begin{abstract}
We propose simple and flexible training and decoding methods for influencing output style and topic in neural encoder-decoder based language generation.
This capability is desirable in a variety of applications, including conversational systems, where successful agents need to produce language in a specific style and generate responses steered by a human puppeteer or external knowledge. 
We decompose the neural generation process into empirically easier sub-problems: a faithfulness model and a decoding method based on selective-sampling. 
We also describe training and sampling algorithms that bias the generation process with a specific language style restriction, or a topic restriction.
Human evaluation results show that our proposed methods are able to restrict style and topic without degrading output quality in conversational tasks.
\end{abstract}

\section{Introduction}
Neural encoder-decoder models have demonstrated great promise in many sequence generation tasks, 
	including neural machine translation~\cite{Sutskever:14-S2S,Cho:2014-EMNLP,Bahdanau:14-attention,Luong:15-attention,Google-NMT:16}, 
	image captioning~\cite{Xu2015show}, 
	summarization~\cite{Rush:16-summ,Gu:16-copy,graham:16-length},
	and conversation generation~\cite{Vinyals:15-chat,Sordoni:NAACL-2015,Shang:15-respond,Shao:17-long-chat,Jiwei:15-MMI}.
These encouraging early successes have motivated research interest in training more natural-sounding conversational systems 
	based on large volumes of open-domain human-to-human interactions. 
In order to create more human-like patterns of conversation, the agents need to have recognizable (and tunable) style, just as individual humans do, 
	and also need to accept guidance from separate information processing modules in order to increase quality of responses. 
In an extreme case, an agent may be micro-managed by a human user who uses the neural model to enforce grammar and style (e.g., a level of politeness, or a type of humor), while driving the content directly (e.g., by expressing 
In this manner, the neural model becomes an authoring tool, rather than an independent chat-bot. 
On the other hand, in fully automated agent systems, the agent may be influenced by a knowledge database, or some other artificial information system, while running in a pre-set style or a style deemed best based on the course of the conversation.

\begin{table}[t]
\setlength{\tabcolsep}{4pt}
\center{}
{\small
\begin{tabularx}{\linewidth}{rl}
	\toprule
	                    \ChatContext & Where are you?                                            \\\midrule
	  {{\it Twitter bot}}\xspace & i'm on my way to london now.                              \\
	  {{\it Star Wars bot}}\xspace & we're heading to the dark side of jabba's        \\
                                   & palace. \\
	  {{\it Hillary bot}}\xspace & i'm here in philadelphia, pennsylvania                    \\
	    {{\it Trump bot}}\xspace & i went to the white house. we were in           \\
                                 & brooklyn.                                                  \\
	  {{\it Kennedy bot}}\xspace & i am on the question of the united states.                \\
      
      \hline \midrule
	                    \ChatContext & What are you doing?                                       \\\midrule
	  {{\it Twitter bot}}\xspace & i'm just chilling with my dad, i'm just            \\
	                             & sitting in the kitchen and listening to music                     \\
	  {{\it Star Wars bot}}\xspace & i'm doing my magic solo.                                \\
	  {{\it Hillary bot}}\xspace & i'm running for president, i'm going to be                \\
	                             & talking about some of these things                        \\
	    {{\it Trump bot}}\xspace & i'm building a wall, right now                            \\
	  {{\it Kennedy bot}}\xspace & today, i am asking the congress for a new                 \\
	                             & program to make a new effort to increase               \\
	                             & the tax privileges and to stimulate                           \\ 
      \hline
\end{tabularx}
}
\caption{Example responses from our Star Wars, Hillary, Trump, and Kennedy bots with scented conversation models.}
\label{sample}
\end{table}

One obstacle to achieving this with neural language generation models is that the sentence representation is distributed across all coordinates of the embedding vector in a way that is hard to disentangle, and thus control. 
In order to gain insight into the full distribution of what a decoder might produce given the prompt sentence as input, the model has to be heavily (and sometimes cleverly) sampled. The second problem is that neural models only become highly functional after training with very large amounts of data, while the strongly recognizable style usually must be defined by a relatively tiny corpus of examples (e.g., all Seinfeld episodes, or all popular song lyrics).

In this paper, we address the challenge of how to enforce the decoder models to mimic a specific language style with only thousands of target sentences, as well as generating specific content in that style. 
We developed and experimented with several training and decoding procedures to allow the model to adapt to target language style and follow additional content guidance. 
Our experiments, conducted on an open-domain corpus of Twitter conversations and small persona corpora, show that our methods are capable of responding to queries in a transferred style without significant loss of relevance, and can respond within a specific topic as restricted by a human. Some examples of `scenting' the base conversation model with particular styles are shown in Table \ref{sample}. 
More can be found in the Supplementary Material.

\section{Related Work}

Recurrent neural network based encoder-decoder models have been applied 
	to machine translation and quickly achieved state-of-the-art 
    results \cite{Bahdanau:14-attention, Luong:15-attention}.
As an extension, the attention mechanism enables the decoder to revisit the input 
	sequence's hidden states and dynamically collects information needed for each decoding step.
Specifically, our conversation model is established based on a combination of the models of 
	\cite{Bahdanau:14-attention} and \cite{Luong:15-attention} that we found to be effective.
In section \ref{sec:s2s}, we describe the attention-based neural encoder-decoder model we used in detail.

This work follows the line of research initiated by \cite{Ritter:11} and \cite{Vinyals:15-chat} who treat generation of conversational dialog as a data-drive statistical machine translation (SMT) problem.
\citet{Sordoni:NAACL-2015} extended \cite{Ritter:11} by re-scoring SMT outputs using a neural encoder-decoder model conditioned on conversation history.
Recently, researchers have used neural encoder-decoder models to directly generate responses in an end-to-end fashion without relying on SMT phrase tables\cite{Vinyals:15-chat,Sordoni:NAACL-2015,Shang:15-respond,Shao:17-long-chat,Jiwei:15-MMI}.

\citet{Jiwei:16-persona} defined a ``persona'' as the character that an artificial agent, as actor, plays or performs during conversational interactions.
Their dataset requires user identification for all speakers in the training set, while our methods treat the base data (millions of twitter conversations) as unlabeled, and the target persona is defined simply by a relatively small sample of their speech. In this sense, the persona can be any set of text data. In our experiments, for example, we used a generic Star Wars character that was based on the entire set of Star Wars scripts (in addition to 46 million base conversations from Twitter). This provides us with a system that can talk about almost anything, being able to respond to most prompts, but in a recognizable Star Wars style. Other possibilities include training (styling) on famous personalities, or certain types of poetry, or song lyrics, or even mixing styles by providing two or more datasets for styling. Thus our targets are highly recognizable styles, and use of these for emphasis (or caricature) by human puppeteers who can  choose from multiple options and guide neural models in a direction they like. We expect that these tools might not only be useful in conversational systems, but could also be popular in social media for text authoring that goes well beyond spelling/grammar auto correction.

\section{Neural Encoder-Decoder Background}
\label{sec:s2s}
In general, neural encoder-decoder models aim at generating a target sequence $Y = \left( y_1, \ldots, y_{T_y} \right)$
	given a source sequence $X = \left( x_1, \ldots, x_{T_x} \right)$.
Each word in both source and target sentences, $x_t$ or $y_t$, 
	belongs to the source vocabulary $V_x$, and the target vocabulary $V_y$ respectively.

First, an encoder converts the source sequence $X$ into
a set of context vectors $C = \left\{ \vh_1, \vh_2, \ldots, \vh_{T_x} \right\}$,
whose size varies with regard to the length of the source passage.  
This context representation is generated using a multi-layered recurrent neural network (RNN).
The encoder RNN reads the source passage from the first token until the last one, where
${\vh}_i = {\Psi}\left( {\vh}_{i-1}, \mE_x\left[x_t\right] \right).$
Here $\mE_x \in \RR^{|V_x| \times d}$ is an embedding matrix containing vector representations of words,
and ${\Psi}$ is a recurrent activation unit that we employ in the Long Short-Term Memory (LSTM) \cite{Hochreiter:1997-LSTM}. 

The decoder, which is also implemented as an RNN, generates one word at a time, 
	based on the context vector set returned by the encoder. 
The decoder's hidden state $\bar{\vh}_t$ is a fixed-length continuous vector that is updated in the same way as encoder.
At each time step $t$ in the decoder, a time-dependent attentional context vector $\vc_t$ 
	is computed based on the current hidden state of the decoder
	$\bar{\vh_t}$ and the whole context set $C$. 

Decoding starts by computing the content-based score of each context vector as:
    $e_{t, i} = \bar{\vh}^\top_t W_a \vh_i.$
This relevance score measures how helpful the $i$-th
context vector of the source sequence is in predicting next word based on the decoder's current hidden state $\bar{\vh}^\top_t$.
Relevance scores are further normalized by the softmax function: $\alpha_{t, i} = \frac{\exp(e_{t,i})}{\sum_{j=1}^{T_x} \exp(e_{t, j})},$
and we call $\alpha_{t, i}$ the attention weight.
The time-dependent context vector $\vc_t$ is then the weighted sum of the
context vectors with their attention weights from above: $\vc_t = \sum_{i=1}^{T_x} \alpha_{t, i} \vh_i$.

With the context vector $\vc_t$ and the hidden state
$\vh_{t}$, we then combine the information from both vectors to produce an attentional hidden state as follow:
	$\vz_t = \tanh(\mW_c[\vc_t; {\vh}_t])$.
The probability distribution for the next target symbol is computed by
	$p(y_t = k|\tilde{y}_{< t}, X) \propto \exp(\mW_s\vz_t + \vb_t )$.

\section{Decoding with Selective Sampling}
\label{sec:selective}
The standard objective function for neural encoder-decoder models is the log-likelihood of target $T$ given source $S$, which at test time yields the statistical decision problem:
\begin{equation}
\hat{T}= \argmax_T \big\{\log p(T|S)\}.
\label{pts}
\end{equation}
However, as discussed in \cite{Jiwei:15-MMI,Shao:17-long-chat}, simply conducting beam search over the above objective will tend to generate generic and safe responses that lack diversity, such as \textit{``I am not sure''}. 
In section \ref{sec:rank}, we present a ranking experiment in which we verify that an RNN-based neural decoder provides a poor approximation of the above conditional probability, and instead biases towards the  target language model $p(T)$. 
Fortunately, the backward model $p(S|T)$ empirically perform much better than $p(T|S)$ on the relevance ranking task.
Therefore, we directly apply Bayes' rule to Equation \ref{pts}, as in statistical machine translation \cite{Brown:93-SMT}, and use: 
\begin{align}
  \hat{T} &= \argmax_T \big\{\log p(S|T) + \log p(T) \}.
  \label{pst}
\end{align}
Since $p(T|S)$ is empirically biased towards $p(T)$, in practice, this objective also resembles the Maximum Mutual Information (MMI) objective function in \cite{Jiwei:15-MMI}.

The challenge now is to develop an effective search algorithm for a target words sequence that maximize the product in Equation \ref{pst}.
Here, we follow a similar process as in \cite{wen:15-dialog} which generates multiple target hypotheses with stochastic sampling based on $p(T|S)$, and then ranks them with the objective function \ref{pst} above.
However, as also observed by \cite{Shao:17-long-chat}, step-by-step naive sampling can accumulate errors as the sequence gets longer.

To reduce language errors of stochastic sampling, we introduce a sample selector to choose the next token among $N$ stochastically sampled tokens based on the predicted output word distributions.
The sample selector, which is a multilayer perceptron in our experiments, takes the following features: 
1) the log-probability of current sample word in $p(w_t|S)$;
2) the entropy of current predicted word distribution, 
	$\sum_{w_t} P(w_t|S) \log P(w_t|S)$ for all $w_t$ in the vocabulary;
3) the log-probability of current sample word in $p(w_t|\emptyset)$, which we found effective in ranking task. 
The selector outputs a binary variable that indicates whether the current sample should be accepted or rejected.

At test time, if none of the $N$ sampled tokens are above the classification threshold, we choose the highest scored token.
If there are more than 1 acceptable samples among $N$ stochastically sampled tokens, we randomly choose one among them.
Ideally, this permits us to safely inject diversity while maintaining language fluency.
We also use the sample acceptor's probabilities as the language model score $P(T)$ for objective in equation \ref{pst}.

As regards directly integrating beam-search, we found (a) that beam-search often produces a set of similar top-N candidates, and (b) that decoding with only the objective $p(Y|X)$ can easily lead to irrelevant candidates. (See section \ref{sec:rank}) 
Therefore, we use the selective-sampling method to generate candidates for all our experiments; this (a) samples stochastically then (b) selects using a learned objective from data. 
The sample-then-select approach encourages more diversity (v.s. MMI's beam-search) while still maintain language fluency (v.s. naive-sampling).

\section{Output style restriction using a small `scenting' dataset}

In this section, we propose three simple yet effective methods of 
	influencing the language style of the output in the neural encoder-decoder framework.
Our language style restricting setup assumes that there is a large open-domain parallel corpus
	that provides training for context-response relevance, 
	and a smaller monologue speaker corpus that reflects the language characteristics of the target speaker. We will refer to this smaller set as a `scenting' dataset, since it hints at, or insinuates, the characteristics of the target speaker.

\subsection{$Rank$: Search in the Target Corpus}
\label{sec:rankall}
Our first approach to scenting is to simply use the all sentences in the target speaker's corpus as generation candidates, ranked by the objective (\ref{pst}) for a given prompt.
Since these sentences are naturally-occurring instead of generated word-by-word, we can safely assume $p(T)$ is constant (and high), and so the objective only requires sorting the sentences based on the backward model $p(S|T)$.

RNN-based ranking methods are among the most effective methods 
	for retrieving relevant responses \cite{Di:2015-ACL,Di:2016-TREC}. Thus this approach is a very strong baseline. 
Its limitation is also obvious: by limiting all possible responses to a fixed finite set of sentences, this method 
	cannot provide a good response if such a response is not already in the scenting dataset.

\subsection{$Multiply$: Mixing the base model and the target language model during generation}
\label{sec:mul}
In our second method we use both the vanilla encoder-decoder model trained on open-domain corpus 
	and the target domain language model trained on the corpus while decoding output sentence.
The idea is to use a speaker's language model, which is also RNN-based in our experiments,
	to restrict the open-domain encoder-decoder model's step-by-step word prediction.
Similar ideas have been tested in domain adaptation for statistical machine translation \cite{Koehn:07}, 
 	where both in-domain and open-domain translation tables were used as candidates for generating target sentence.
Because open-domain encoder-decoder models are trained with various kinds of language patterns and topics, 
	choosing a sequence that satisfies both models may produce relevant responses that are also in the target language style.  
We found that a straightforward way of achieving this is to multiply the two models' distributions 
	$p_1(t|S)^{\lambda_1}p_{2}(t)^{\lambda_2}$ at each point and then re-normalize before sampling.
The weights can be tuned either by the perplexity on the validation set, or through manually controlling the trade-off between style restriction and answer accuracy.

\subsection{$Finetune$:  Over-training on Target Corpus with Pseudo Context}
\label{sec:finetune}
Fine-tuning is a widely used in the neural network community to achieve transfer learning.
This strategy permits us to train the neural encoder-decoder on a larger general parallel corpus, and then use the learned parameters to initialize the training of a styled model.
Most of the time, however, the target speaker's corpus will lack training data in parallel form. 
For example, if we train on song lyrics or movie scripts, or political speeches, the data will not be in a question-answer form.
To make encoder-decoder overtraining possible, we treat every sentence in the scenting corpus as a target sentence $T$ generated a pseudo context from the backward model $p(S|T)$ trained on the open-domain corpus.
Over-training on such pairs imparts the scenting dataset's language characteristics, while retaining the generality of the original model.
We also found that the previous sentence in the styled corpus (i.e., previous sentence in the speech) provides helpful context for the current sentence, analogous with a question-answer link.
Thus we use both pseudo context and the previous sentence as possible sources $S$ to fine-tune the in-domain decoder.
To avoid overfitting, we stop overtraining when the perplexity on the in-domain validation set starts to increase.
A corresponding sample acceptor is also trained for the fine-tuned model: we found it helpful to initialize this from the open-domain model's sample acceptor.

\section{Restricting the Output Topic}

We further introduce a topic restricting method for neural decoders based on the Counting Grid \cite{Jojic:11-UAI-CG} model, by treating language guidance as a topic embedding.
Our model extension provides information about the output topic in the form of an additional topic embedding vector to the neural net at each time step.

\subsection{$CG$: Counting Grids} 
\label{sec:cg}

The basic counting grid $\pi_{\bk}$  is a set of distributions on the $d$-dimensional toroidal discrete grid $\bE$ indexed by $\bk$. The grids in this paper are bi-dimensional and typically from $(E_x = 32) \times (E_y =32)$ to $(E_x = 64) \times (E_y = 64)$ in size. The index $z$ indexes a particular word in the vocabulary $z=[1\dots Z]$. Thus, $\pi_{\bi}(z)$ is the probability of the word $z$ at the $d$-dimensional discrete location $\bi$, and $\sum_z \pi_{\bi}(z)=1$ at every location on the grid. The model generates bags of words, each represented by a list of words $\bw = \{ w_n \}_{n=1}^N$ with each word $w_n$ taking an integer value between $1$ and $Z$.
The modeling assumption in the basic CG model is that each bag is generated from the distributions in a single window $\bW$ of a preset size, e.g., $(W_x = 5)\times (W_y=5)$. 
A bag can be generated by first picking a window at a $d$-dimensional location $\ell$, denoted as $W_{\ell}$, then generating each of the $N$ words by sampling a location $\bk_n$  for a particular micro-topic $\pi_{\bk_n}$ uniformly within the window, and sampling from that micro-topic.

Because the conditional distribution $p(\bk_n|\ell)$ is a preset uniform distribution over the grid locations inside the window placed at location $\ell$, the variable $\bk_n$ can be summed out \cite{Jojic:11-UAI-CG}, and the generation can directly use the grouped histograms
{%
\begin{equation} \label{eq:h} 
h_{\ell}(z)=\frac{1}{|\bW|} \sum_{\bj \in W_{\ell}} \pi_{\bj}(z),
\end{equation}
}
where $|\bW|$ is the area of the window, e.g. 25 when 5$\times$5 windows are used. 
In other words, the position of the window $\ell$ in the grid is a latent variable given which we can write the probability of the bag as
{
\begin{equation} 
\small
\label{eq:cg}
P( \bw | \ell )= \prod_{w_n \in \bw} h_{\ell}(w_n) =\prod_{w_n \in \bw} \big( \frac{1}{|\bW|} \cdot \sum_{\bj \in W_\ell} \pi_{\bj}(w_n) \big)
\end{equation}
}
As the grid is toroidal, a window can start at any position and there is as many $h$ distributions as there are $\pi$ distributions. The former will have a considerably higher entropy as they are averages of many $\pi$ distributions.
Although the basic CG model is essentially a simple mixture assuming the existence of a single source (one window) for all the features in one bag, it can have a very large number of (highly related) choices $h$ to choose from. Topic models \cite{Blei:2003-LDA,Lafferty:06-CTM}, on the other hand, are admixtures that capture word co-occurrence statistics by using a much smaller number of topics that can be more freely combined to explain a single document (and this makes it harder to visualize the topics and pinpoint the right combination of topics to use in influencing the output).

In a well-fit CG model, each data point tends to have a rather peaky posterior location distribution because the model is a mixture.   
The CG model can be learned efficiently using the EM algorithm because the inference of the hidden variables, as well as updates of $\pi$ and $h$ can be performed using summed area tables \cite{Crow:1984}, and are thus considerably faster than most of the sophisticated sampling procedures used to train other topic models. The use of overlapping windows helps both in controlling the capacity of the model and in organizing topics on the grid automatically: Two overlapping windows have only slightly different $h$ distributions, making CGs especially useful in visualization applications where the grid is shown in terms of the most likely words in the component distributions $\pi$ \cite{Perina2014}.\footnote{\cite{XingWWLHZM17} have recently proposed using LDA for topic modeling in Sequence-To-Sequence response generation models. We believe that the CG embedding used here will prove easier to apply and interpret through visualization.} 

\begin{figure*}
  \centering
  \includegraphics[width=\textwidth]{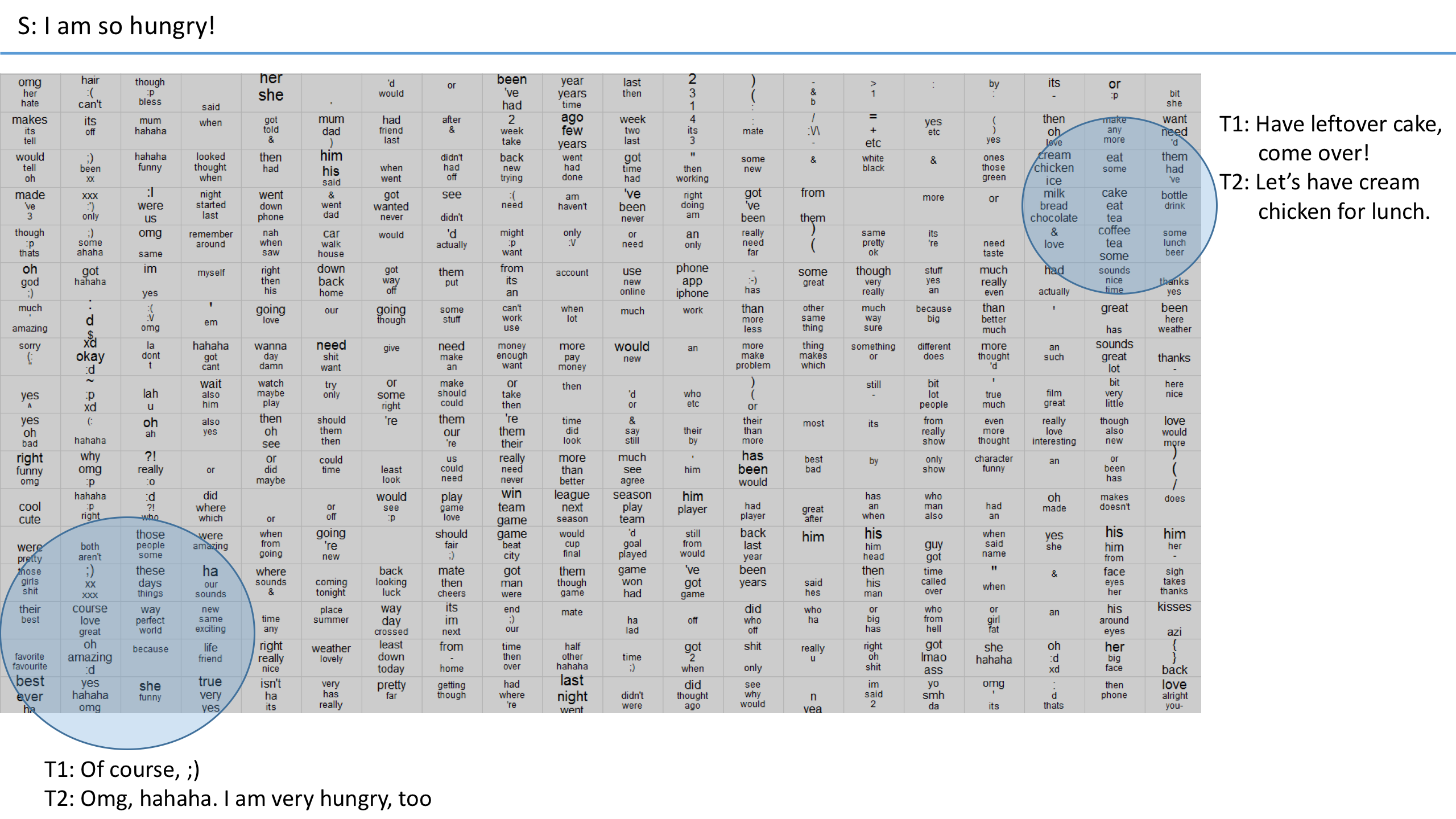}
  \caption{A part of a Counting Grid trained on Twitter data and its use in providing topical hints in decoding. For the source sentence at the top, the decoder may produce the two target samples on the right, if the circled locations are used as a hint, or the two sentences at the bottom if the locations in the lower right are picked.}
  \label{fig:cg}
\end{figure*}

Having trained the grid on some corpus (in our case a sample of the base model's corpus), the mapping of either a source $S$ and/or target $T$ sentence can be obtained by treating the sentences as bags of words. 
%%%fix
By appending one or both of these mappings to the decoder's embedding of the target $T$, the end-to-end encoder-decoder learning can be performed in a scenario where the decoder is expected to get an additional hint through a CG mapping. 
In our experiments, we only used the embedding of the target $T$ as the decoder hint, and we appended the full posterior distribution over CG locations to the encoder's embedding. At test time, we only have the $S$ and need to generate $T$ without knowing where it may map in the counting grid. 
We considered two ways of providing a mapping:
\begin{itemize}[noitemsep,topsep=0pt,parsep=0pt,partopsep=0pt]
\item The user provides a hint sentence $H$ (could be just a few words in any order), and the CG mapping of the user's hint, i.e. the full posterior distribution $p(\ell|H)$, is used in the decoding. 
The posterior probabilities over $32\times 32$ grid locations are unwrapped  into a vector with a size of $|L|=1024$, and then concatenated with the word embedding as the input at each time-step.
That acts to expand the user's hint into a sentence with similar content (and style if the model is also styled).
\item The CG is scanned and a variety of mappings are tested as inputs to provide a diverse set of possible answers. In our experiments, instead of scanning over all 1024 possible locations in the grid, we retrieved several possible answers using information retrieval (ranking of the data samples in the training set based on the source $S$ and picking the top ten). Then the CG mapping $p(\ell|H)$ of these retrieved hints is used to decode several samples from each. 
\end{itemize}
As an example, Figure~\ref{fig:cg} shows a portion of a CG trained on randomly chosen 800k tweets from the twitter corpus. In each cell of the grid, we show the top words in the distribution $\pi_{\bj}(z)$ over words ($z$) in that location ($\bj$). (Each cell has a distribution over the entire vocabulary). As a response to ``I am hungry," using two highlighted areas as hints, we can generate either a set of empathic responses, such as `Me too,' or food suggestions, such as `Let's have cake.' It will also be evident that some areas of the grid may produce less sensical answers. These can later be pruned by likelihood criteria or by user selection. 

\section{Experiments}

\subsection{Datasets}

\paragraph{Yahoo!\ Answer Dataset.}
We use the Comprehensive Questions and Answers dataset\footnote{\url{http://webscope.sandbox.yahoo.com/catalog.php?datatype=l}} 
	to train and validate the performances of different decoding setups with ranking experiments described in section \ref{sec:rank}.
This dataset contains 4.4 million \YA questions and the user-selected best answers. 
Unlike the conversational datasets, such as the Twitter dataset described below, 
 	it contains more relevant and specific responses for each question, 
 	which leads to less ambiguity in ranking.  
    
\paragraph{Twitter Conversation Dataset.} 
We trained our base encoder-decoder models on the Twitter Conversation Triple Dataset described in \cite{Sordoni:NAACL-2015},
	which consists of 23 million conversational snippets randomly selected from a collection of 129M
	context-message-response triples extracted from the Twitter Firehose over the 3-month period from June through August 2012.
For the purposes of our experiments, we split the triples into context-message and message-response pairs yielding 46M source-target pairs.
For tuning and evaluation, we used the development dataset of size 200K conversation pairs and the test dataset of 5K examples.
The corpus is preprocessed using a Twitter specific tokenizer \cite{OConnor:10-tweet}. 
The vocabulary size is limited to 50,000 
excluding the special boundary symbol and the unknown word tag.

\paragraph{Scenting datasets.}
A variety of persona characters have been trained and tested, including Hillary Clinton, Donald Trump, John F. Kennedy, Richard Nixon, singer-songwriters, stand-up comedians, and a generic Star Wars character.
In experiments, we evaluated on a diverse set of representative target speakers:

\textbf{JFK.} 
We mainly tested our models on John F. Kennedy's speeches collected from American Presidency Project\footnote{\url{http://www.presidency.ucsb.edu/}}, 
which contains 6474 training and 719 validation sentences.

\textbf{Star Wars.} 
Movie subtitles of three Star Wars movies are also tested\footnote{
\citet{rewriting-algebra-16} also uses Star Wars scripts to test theme rewriting of algebra word problems.}. 
They are extracted from Cornell Movie-Dialogs Corpus \cite{Danescu:11a}, and
have 495 training and 54 validation sentences.

\textbf{Singer-Songwriter.} 
We also evaluated our approach on a lyric corpus from a collective of singers: Coldplay, Linkin Park, and Green Day.
The lyric dataset is collected from mldb.org and has 9182 training and 1020 validation lines.

\paragraph{Debate Chat Contexts.}
We designed testing questionnaires with 64 chat contexts spanning a range of topics in politic, science, and technology: 
	the sort of questions we might ask in an entertaining political debate.\footnote{See the Supplementary material.}
To test the model's ability to control output topic in section \ref{sec:exp-topic}, we also created one hint per question. 

\subsection{Network Setup and Implementation}
\label{sec:impl}
Our encoder and decoder RNNs contains two-layer stacked LSTMs. 
Each LSTM layer has a memory size of 500.
The network weights are randomly initialized using a uniform distribution ($-0.08, 0.08$), 
	and are trained with the ADAM optimizer \cite{Kingma:14-adam}, 
	with an initial learning rate of 0.002.
Gradients were clipped so their norm does not exceed 5.
Each mini-batch contains 200 answers and their questions.
The words of input sentences were first converted to 300-dimensional vector representations 
	learned from the RNN based language modeling tool word2vec \cite{mikolov2013distributed}.
The beginning and end of each passage are also padded with a special boundary symbol.
During decoding, our model generates 500 candidate samples in parallel, then ranks them.
As these are processed in batches on GPU, generation is very efficient. 
We also experimented incorporating an information retrieval (IR) module to automatically collect topic hints for CG-based decoder.
Specifically, a full-text index of twitter corpus is built using solr\footnote{\url{https://lucene.apache.org/solr/}},
and the top 10 searched results based on the source sentence are be used to generate posterior CG distributions as hints. 

\subsection{Validating the Decoding Setup with Ranking}
\label{sec:rank}
We performed a ranking evaluation applying different decoding setups on the \YA dataset.
Here we wanted to test the relevance judgment capacities of different setups, 
	and validate the necessity of the new decoding method discussed in section \ref{sec:selective}.
\YA question is used as source $S$, and its answer is treated as target $T$.
Each test question is associated with one true answer and 19 random answers from the test set.
MRR (Mean Reciprocal Rank) and P@1 (precision of top1) were then used as evaluation metrics.

Table \ref{table:rank} shows the answer ranking evaluation results: 
	the forward model $P(T|S)$, by itself is close to the performance of random selection in  
	distinguishing true answer from wrong answers.
This implies that a naive beam search over only the forward model may generate irrelevant outputs.	
One hypothesis was that $P(T|S)$ is biased toward $P(T)$, and performance indeed improves after normalizing by $P(T)$.
However, it is difficult to directly decode with objective $ P(T|S)/P(T|\emptyset)$, 
	because this objective removes the influence of the target-side language model.
Decoding only according to this function will thus result in only low-frequency words and ungrammatical sentences, 
	behavior also noted by \citep{Jiwei:15-MMI,Shao:17-long-chat}. 

\begin{table}[h]
  \centering
  \hspace*{-0.5cm}
  \begin{tabular}{l|l|l}
    \toprule
  	\textbf{Ranking Methods}             & \textbf{MRR} & \textbf{P@1} \\ \hline\hline
  	$ P_{rnn}(T|S)$                      & 0.224        & 0.075        \\ \hline
  	$ P_{rnn}(T|S)/P_{rnn}(T|\emptyset)$ & 0.652        & 0.524        \\ \hline
  	$ P_{rnn}(S|T)$                      & 0.687        & 0.556        \\ \hline
  \end{tabular}
  \caption{Ranking the true target answer among random answers on \YA test set.}
  \label{table:rank}
\end{table}

\subsection{Human Evaluations}

\subsubsection{Systems}

We tested 10 different system configurations to evaluate the overall output quality, 
	and their abilities of influencing output language style and topic:
\begin{itemize}[noitemsep,topsep=0pt,parsep=0pt,partopsep=0pt]
	\item \textbf{\NS} each word in the target. 
	\item \textbf{\Sel} as described in section \ref{sec:selective}; all the following systems are using it as well.
	\item \textbf{\CgIR} uses IR results to create counting grid topic hints (sections \ref{sec:cg} and \ref{sec:impl}).
	\item \textbf{\Rank} uses proposals from the full JFK corpus as in section \ref{sec:rankall}.
	\item \textbf{\Multiply} with a JFK language model as in section \ref{sec:mul}.
	\item \textbf{\Finetune} with JFK dataset as in section \ref{sec:finetune}.
	\item \textbf{\FinetuneCgIR} uses IR results as topic hints for fine-tuned JFK.
	\item \textbf{\FinetuneCgTopic} forced to use the given topic hint for fine-tuned JFK.
	\item \textbf{\Lyric} fine-tuned cg-topic.
	\item \textbf{\StarWar} fine-tuned cg-topic.
\end{itemize}

\subsubsection{Evaluation Setup}
Owing to the low consistency between automatic metrics and human perception on conversational tasks \cite{Liu:16-no-eval,Stent:05eval} 
	and the lack of true reference responses from persona models,
	we evaluated the quality of our generated text with a set of judges recruited from Amazon Mechanical Turk (AMT). 
Workers were selected based on their AMT prior approval rate ($>$95\%).
Each questionnaire was presented to 3 different workers.
We evaluated our proposed models on the 64 debate chat contexts. 
Each of the evaluated methods generated 3 samples for every chat context.
To ensure calibrated ratings between systems, we show the human judges all system outputs (randomly ordered) for each particular test case at the same time. 
For each chat context, we conducted three kinds of assessments:

\paragraph*{Quality Assessment}
Workers were provided with the following guidelines: 
	``Given the chat context, a chat-bot needs to continue the conversation.
		Rate the potential answers based on your own preference on a scale of 1 to 5 (the highest):''
\begin{itemize}[noitemsep,topsep=0pt,parsep=0pt,partopsep=0pt]
	\item 5-Excellent: ``Very appropriate response, and coherent with the chat context.''
	\item 4-Good: ``Coherent with the chat context.''
	\item 3-Fair: ``Interpretable and related. It is OK for you to receive this chat response.''
	\item 2-Poor: ``Interpretable, but not related.''
	\item 1-Bad:  ``Not interpretable.''
\end{itemize}
In this test, the outputs of all 10 systems evaluated are then provided to worker together for a total of 30 responses.
In total, we gathered $64\cdot30\cdot3=5760$ ratings for quality assessments, and 47 different workers participated.

\paragraph{Style Assessment.}
We provided following instructions:
``Which candidate responses are likely to have come from or are related to [Persona Name]?''. 
Checkboxes were provided for the responses from style-influenced systems and from {\Sel} as a baseline.

\paragraph*{Topic  Assessment.}
The instruction was:
``Which candidate answers to the chat context above are similar or related to the following answer: `[a hint topic provided by us]'?''.
This was also a checkbox questionnaire. Candidates are from both style- and topic-influenced systems (fine-tuned cg-topic), and from {\Sel}  as a baseline.

\subsubsection{Results}
\begin{table}[t]
	\centering
	\begin{tabularx}{\linewidth}{l|l|l}
        \toprule
		\textbf{Methods}           & \textbf{Quality (MOS)} & \textbf{Style}   \\ 		      \hline\hline
		\NS                        & 2.286 $\pm$ 0.046        & \NA               \\ \hline
		\Sel                       & 2.681 $\pm$ 0.049        & 10.42\%            \\ \hline
		\CgIR                      & 2.566 $\pm$ 0.048        & 10.24\%            \\ \hline
		\Rank                      & 2.477 $\pm$ 0.048        & 21.88\%            \\ \hline
		\Multiply                  & 2.627 $\pm$ 0.048        & 13.54\%            \\ \hline
		\Finetune                  & 2.597 $\pm$ 0.046        & 20.83\%            \\ \hline
		\FinetuneCgIR              & 2.627 $\pm$ 0.049        & 20.31\%            \\ \hline
		\FinetuneCgTopic           & 2.667 $\pm$ 0.045        & 21.09\%            \\ \hline
		\Lyric                     & 2.373 $\pm$ 0.045        & \NA                \\ \hline
		\StarWar                   & 2.677 $\pm$ 0.048        & \NA                \\ \hline
	\end{tabularx}
	\caption{\label{table:mos} Results of quality assessments with 5-scale mean opinion scores (MOS) and JFK style assessments with binary ratings.
		Style results are statistically significant compared to the {\Sel} by paired t-tests ($p<0.5\%$).}
\end{table}

\paragraph*{Overall Quality.}
We conducted mean opinion score (MOS) tests for overall quality assessment of generated responses with questionnaires described above.
Table \ref{table:mos} shows the MOS results with standard error.
It can be seen that all the systems based on selective sampling are significantly better than vanilla sampling baseline.
When restricting output's style and/or topic, the MOS score results of most systems do not decline significantly except {\Lyric}, 
	which attempts to generate lyrics-like outputs in response to to political debate questions, resulting in uninterpretable strings.
    
Our {\Rank} method uses $p(S|T)$ to pick the answer from the original persona corpus, and is thus as good at styling as the person themselves. 
Because most of our testing questionnaire is political, the {\Rank} was indeed often able to find related answers in the dataset (JFK). 
Also, unlike generation-based approaches, {\Rank} has oracle-level language fluency and it is expected to have quality score of at least 2 (``Interpretable, but not related''). 
Overall, however, the quality score of {\Rank} is still lower than other approaches. 
Note that a hybrid system can actually chose between rank and the decoder's outputs based on likelihood, as shown in the example of bJFk-bNixon debate in the supplemental material.

\paragraph*{Influencing the Style.}

Table \ref{table:mos} also shows the likelihood of being labeled as JFK for different methods. 
It is encouraging that \textit{finetune} based approaches have similar chances as the {\Rank} system which retrieves sentences directly from JFK corpus, and are significantly better than the {\Sel} baseline. 

\paragraph{Influencing both Style and Topic.}
\label{sec:exp-topic}

\begin{table}[t]
\centering
\hspace*{-0.25cm}{
\begin{tabular}{l|l|l|l|l}
	\toprule
	\multicolumn{1}{c|}{\multirow{2}{*}{\textbf{Persona}}} & \multicolumn{2}{c|}{\textbf{Style}} & \multicolumn{2}{c}{\textbf{Topic}} \\ \cline{2-5}
	\multicolumn{1}{c|}{}                         & Ours & Base                & Ours & Base               \\ \hline\hline
	John F. Kennedy                               & 21\% & 10\%                & 33\% & 22\%               \\ \hline
	Star Wars                                     & 27\% & 3\%                 & 14\% & 8\%                \\ \hline
	Singer-Songwriter                             & 31\% & 23\%                & 17\% & 9\%                \\ \hline
\end{tabular}
}
\caption{\label{table:style-topic} The style and topic assessments (both binary) of three models with different personas and with restriction of specific target topic for each chat context. All style and topic results are statistically significant compared to the Base (\Sel) by paired t-tests with $p<0.5\%$.}
\end{table}

Table \ref{table:style-topic} summarizes the results in terms of style (the fraction of answers labeled as in-style for the target persona), and topic (the percentage of answers picked as related to the human-provided topic hint text). We used the last three of the ten listed systems, which are both styled and use specific topic hints to generate answers.
These results demonstrate that it is indeed possible to provide simple prompts to a styled model 
	and drive their answers in a desired direction while picking up the style of the persona.
It also shows that the style of some characters is harder to recreate than others. 
For example, workers are more likely to label baseline results as lyrics from a singer-songwriter than lines from Star Wars movies, 
	which might be because lyrics often take significant freedom with structure and grammar. 
We also found that it is harder for Star Wars and Singer-Songwriter bots to follow topic hints than it is for the John F. Kennedy model, 
	largely because the political debate questions we used overlap less with the topics found in the scenting datasets for those two personas.  

\section{Conclusions}
In this study we investigated the possibility of steering the style and content in the output of a neural encoder-decoder model\footnote{The code and testing data are available at\\  \url{https://github.com/digo/steering-response-style-and-topic}}.
We showed that acquisition of highly recognizable styles of famous personalities, characters, or professionals, is achievable, and that it is even possible to allow users to influence the topic direction of conversations. 
The tools described in the paper are not only useful in conversational systems (e.g., chatbots), but can also be useful as authoring tools in social media. 
In the latter case, the social media users might use neural models as consultants to help with crafting responses to any post the user is reading. 
The AMT tests show that these models do indeed provide increased recognizability of the style, without sacrificing quality or relevance. 

\section*{Acknowledgments}
We thank Shashank Srivastava, Donald Brinkman, Michel Galley, and Bill Dolan for useful discussions and encouragement.
Di Wang is supported by the Tencent Fellowship and Yahoo!\ Fellowship, to which we gratefully acknowledge.

\bibliography{emnlp2017}
\bibliographystyle{emnlp_natbib}

\end{document}

%% file: commands.tex
\newcommand{\ChatContext}{{{\it chat context}}\xspace}

\newcommand{\NS}{\textit{vanilla-sampling}}
\newcommand{\Sel}{\textit{selective-sampling}}
\newcommand{\Rank}{\textit{rank}}
\newcommand{\Multiply}{\textit{multiply}}
\newcommand{\Finetune}{\textit{finetune}}
\newcommand{\FinetuneCgIR}{\textit{finetune-cg-ir}}
\newcommand{\FinetuneCgTopic}{\textit{finetune-cg-topic}}
\newcommand{\CgIR}{\textit{cg-ir}}
\newcommand{\StarWar}{\textit{starwars}}	
\newcommand{\Lyric}{\textit{singer-songwriter}}
\newcommand{\NA}{---}

\newcommand{\YA}{Yahoo!\ Answers\xspace}

\newcommand{\bmx}[0]{\begin{bmatrix}}
	\newcommand{\emx}[0]{\end{bmatrix}}

\newcommand{\vect}[1]{\mathbf{#1}}

\newcommand{\matr}[1]{\mathbf{#1}}

\newcommand{\vb}[0]{\vect{b}}
\newcommand{\vc}[0]{\vect{c}}

\newcommand{\vh}[0]{\vect{h}}

\newcommand{\vz}[0]{\vect{z}}

\newcommand{\mW}[0]{\matr{W}}

\newcommand{\mE}[0]{\matr{E}}

\newcommand{\RR}[0]{\mathbb{R}}

\DeclareMathOperator*{\argmax}{arg\,max}

\newcommand{\bi}{{\bf i}}
\newcommand{\bj}{{\bf j}}
\newcommand{\bk}{{\bf k}}

\newcommand{\bw}{{\bf w}}
\newcommand{\bE}{{\bf E}}
\newcommand{\bW}{{\bf W}}
\newcommand{\cut}[1]{}

\usepackage{algorithmic}